\documentclass[11pt,a4paper]{article}
\usepackage[hyperref]{emnlp2018}
\usepackage{times}
\usepackage{latexsym}

\usepackage{algorithm}
\usepackage[noend]{algpseudocode}
\usepackage{pstricks}
\usepackage{amssymb,fge}

\usepackage{url}

\usepackage{tikz}

\usepackage{amsmath}
\usepackage{amssymb}
\usepackage{multirow}
\usepackage{graphicx}
\aclfinalcopy 

\title{Emo2Vec: Learning Generalized Emotion Representation by Multi-task Training}

\author{Peng Xu, Andrea Madotto, Chien-Sheng Wu, Ji Ho Park {\normalfont and} Pascale Fung  \\
  Center for Artificial Intelligence Research (CAiRE) \\
  The Hong Kong University of Science and Technology, Clear Water Bay \\
  {\tt [pxuab,eeandreamad,cwuak,jhpark,pascale]@ust.hk }
  \\}

\date{}

\begin{document}
\maketitle
\begin{abstract}

In this paper, we propose Emo2Vec which encodes emotional semantics into vectors. We train Emo2Vec by multi-task learning six different emotion-related tasks, including emotion/sentiment analysis, sarcasm classification, stress detection, abusive language classification, insult detection, and personality recognition. Our evaluation of Emo2Vec shows that it outperforms existing affect-related representations, such as Sentiment-Specific Word Embedding and DeepMoji embeddings with much smaller training corpora. When concatenated with GloVe, Emo2Vec achieves competitive performances to state-of-the-art results on several tasks using a simple logistic regression classifier. 






\end{abstract}

\section{Introduction}




Recent work on word representation has been focusing on embedding syntactic and semantic information into fixed-sized vectors~\cite{Mikolov2013,Pennington2014} based on the distributional hypothesis,  and have proven to be useful in many natural language tasks~\cite{Collobert2011}. However, despite the rising popularity regarding the use of word embeddings, they often fail to capture the emotional semantics the words convey. For example, the GloVe vector captures the semantic meaning of ``headache", as it is closer to words of ill symptoms like ``fever" and ``toothache", but misses the emotional association that the word carries. The word ``headache" in the sentence ``You are giving me a headache" does not really mean that the speaker will get a headache, but instead implies the negative emotion of the speaker.

To include affective information into the word representation, \newcite{Tang2016} proposed Sentiment-Specific Word Embeddings (SSWE)  which encodes both positive/negative sentiment and syntactic contextual information in a vector space. This work demonstrates the effectiveness of incorporating sentiment labels in a word-level information for sentiment-related tasks compared to other word embeddings.  However, they only focus on binary labels, which weakens their generalization ability on other affect tasks. \newcite{yu2017refining} instead uses emotion lexicons to tune the vector space, which gives them better results. Nevertheless, this method requires human-labeled lexicons and cannot scale to large amounts of data. \newcite{Felbo2017} achieves good results on affect tasks by training a two-layer bidirectional Long Short-Term Memory (bi-LSTM) model, named DeepMoji, to predict emoji of the input document using a huge dataset of 1.2 billions of tweets. However, collecting billions of tweets is expensive and time consuming for researchers.

Furthermore, most works in sentiment and emotion analysis have focused solely on a single task. Nevertheless, as emotion is a complex concept, we believe that all emotion involving situations such as stress, hate speech, sarcasm, and insult, should be included for a deeper understanding of emotion. Thus, one way to achieve this is through a multi-task training framework, as we present here.  


\noindent\textbf{Contributions}: 1) We propose Emo2Vec \footnote{https://github.com/pxuab/emo2vec\_wassa\_paper}  which are word-level representations that encode emotional semantics into fixed-sized, real-valued vectors. 2) We propose to learn Emo2Vec with a multi-task learning framework by including six different emotion-related tasks. 3) Compared to existing affect-related embeddings, Emo2Vec achieves better results on more than ten datasets with much less training data (1.9M vs 1.2B documents). Furthermore, with a simple logistic regression classifier, Emo2Vec reaches competitive performance to state-of-the-art results on several datasets when combined with GloVe. 

\begin{figure}[t]
  \centering
  \includegraphics[width=\linewidth]{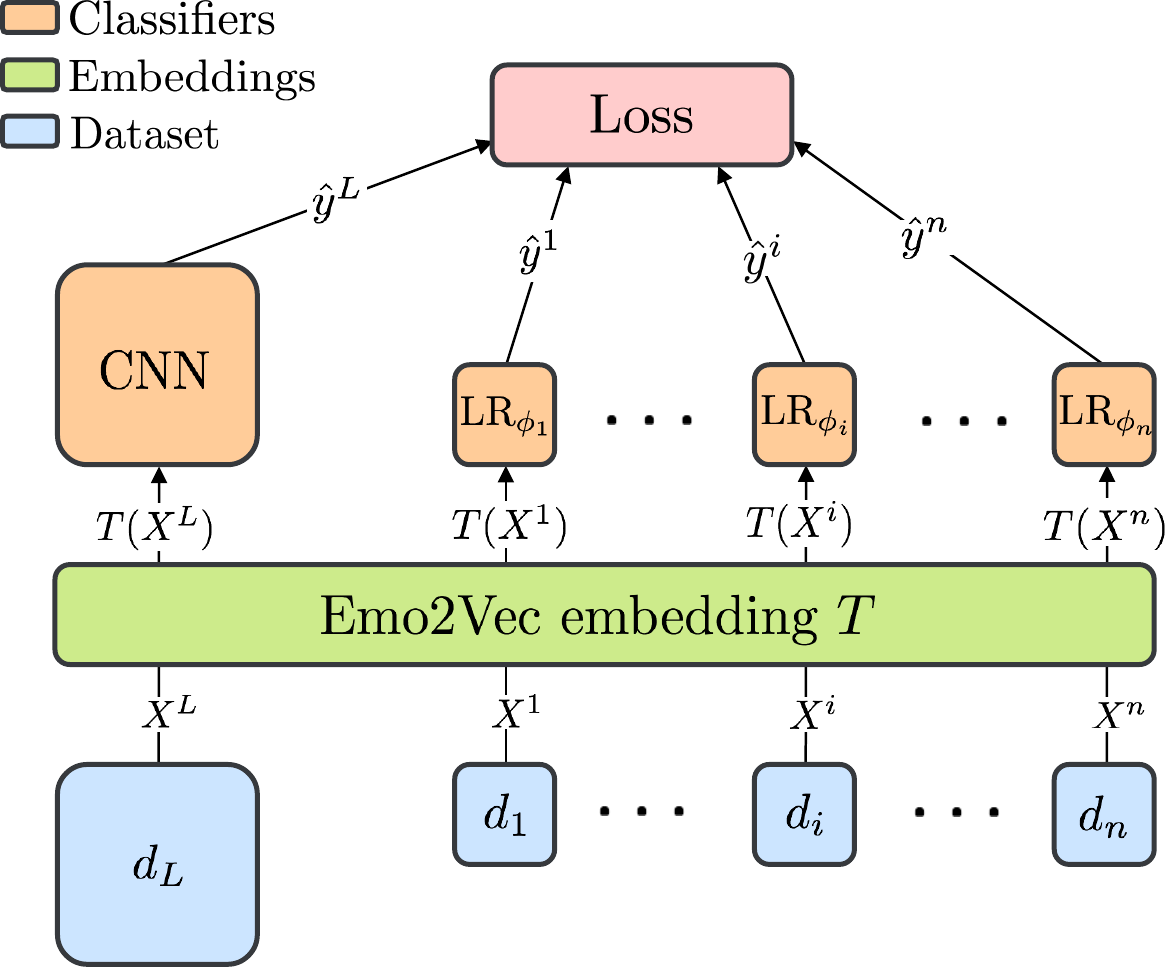}
  \vspace{-10pt}
  
  \caption{Multi-task learning diagram}
  
  \vspace{-10pt}
  \label{fig:multi_task}
\end{figure}

\section{Methodology}
We train Emo2Vec using an end-to-end multi-task learning framework with one larger dataset and several small task-specific datasets. The model is divided into two parts: a shared embedding layer (i.e. Emo2Vec), and task-specific classifiers. All datasets share the same word-level representations (i.e. Emo2Vec), thus forcing the model to encode shared knowledge into a single matrix. For the larger dataset, a Convolutional Neural Network (CNN)~\cite{lecun1998gradient} model is used to capture complex linguistic features present in the corpus. On the other hand, the classifier of each small dataset is a simple logistic regression. 

\paragraph{Notation:} We define $D=\{d_{L},d_1, d_2, \cdots, d_n\}$ as the set of $n+1$ datasets, where $d_L$ is the larger dataset and the other $d_i$ are the small datasets. We denote a sentence $X^i$ with $i \in \{L, 1, 2, \cdots, n\}$ as ${[w_{i,1}, w_{i,2}, \cdots, w_{i,N_i}]}$ where $w_{i,j}$ is the $j$-th word in the $i$-th sample and $N_i$ is the number of words. All the models' parameters are defined as $M_\Phi=\{T, \text{CNN}, \text{LR}_{\phi_1}, \dots, \text{LR}_{\phi_n}\}$, where $T \in \mathbb{R}^{|V|\times k}$ is the Emo2Vec matrix, $|V|$ is the vocabulary size and $k$ is the embedding dimension, CNN is a Convolutional Neural Network model and $LR_{\phi_i}$ for $i \in [1, n]$ is a logistic regression classifier parameterized by $\phi_i $ which is specific for the dataset $d_i$. We denote the embedded representation of a word $w_{i,j}$ with ${e_{w_{i,j}}}$.

\subsection{CNN model}
\begin{figure}[t]
  \centering
  \includegraphics[width=\linewidth]{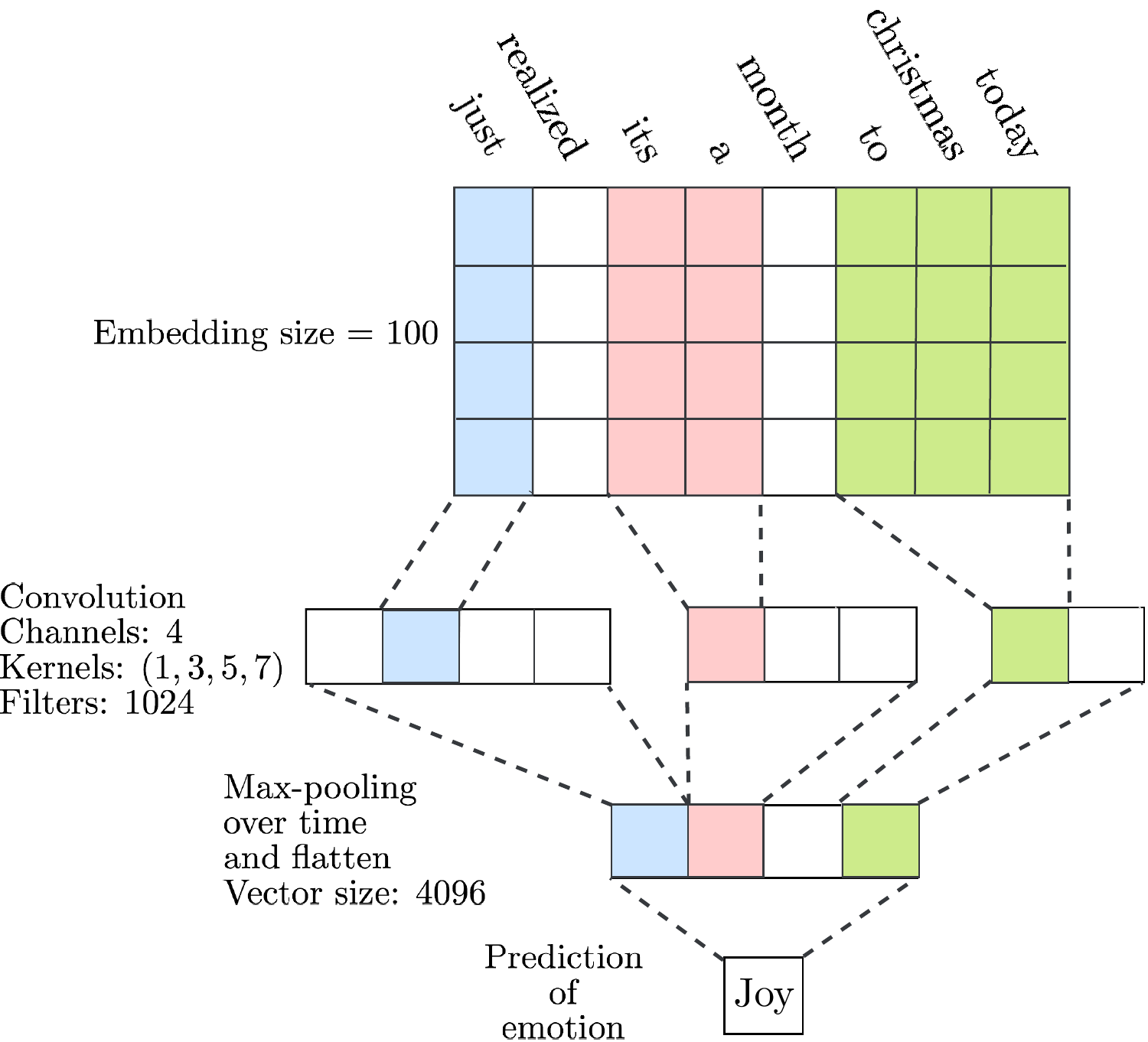}
  \vspace{-10pt}
  
  \caption{Structure of CNN model}
  
  \vspace{-10pt}
  \label{fig:cnn_architecture}
\end{figure}

The CNN architecture used is illustrated in Figure~\ref{fig:cnn_architecture}.  
Firstly, 1-D convolution is used to extract n-gram features from the input embeddings.  Specifically, the $j$-th filter denoted by $F_j$, is convolved with embeddings of words in a sliding window of size $k_j$, giving a feature value $c_{j,t}$. $J$ filters are learned trough this process:
\[c_{j,t} = F_j\,* \, e_{w_{L,t:t+k_j-1}} + b_j\]
where $*$ is the 1-D convolution operation. This is followed by a layer of ReLU activation~\cite{Nair2010} for non-linearity. After that, we add a max-pooling layer of pooling size $M - F_j + 1$ along the time dimension to force the network to find the most relevant feature for predicting $y^L$ correctly. The result of this series of operations is a scalar output of $fm_{j}$. All $fm_{j}$ for $j \in [1, J]$ are then concatenated together to produce a vector representation $fm_{1:J}$ of the whole input sentence.
\[  fm_{j} = \text{Max\_Pooling}\,(\text{ReLU}\,(c_{j,t}))\]

To make the final classification, the vector $fm_{i,1:J}$ is projected to the target label space by adding another fully connected layer (i.e. parameterized by $W$ and $b$), with a softmax activation. 
\[  \hat{y}^L  = 
 \text{Softmax}(W \cdot [fm_{1:J}] + b)\]
 

\subsection{Multi-task learning}

Since collecting a huge amount of labeled datasets is expensive, we collect two types of corpora, one larger dataset (millions of training samples) and a set of small datasets (thousands of training samples each) with accurate labels. For  small datasets, sentiment analysis, emotion classification, sarcasm detection, abusive language classification, stress detection, insult classification and personality recognition are included. The reason why we include many datasets is to 1) leverage different aspects of words emotion knowledge, which may not be present in single domain dataset; 2) create a more general embedding emotional space that can generalize well  across different tasks and domains. To avoid over-fitting, L2 regularization penalty is added from the weights of all logistic regression classifiers $\phi_i$ for $i \in [1, n]$. Hence, we jointly optimize the following loss function:
\[ L(M_\Phi) = \frac{1}{n}\sum_{j=1}^n  L_j +  \lambda\sum_{j=1}^n \| \text{LR}_{\phi_j} \|_2 \]
Where $L_j$ is the negative log likelihood (NLL) between $\hat{y}^j$ and $y^j$, and $\lambda$ an hyper-parameter for the regularization terms. 

\section{Experimental Setup}
\subsection{Dataset}

\subsubsection*{Larger dataset} We collect a larger dataset from Twitter with hashtags as distant supervision. Such distant supervision method using hashtags has already been proved to provide reasonably relevant emotion labels by previous works~\cite{Wang2012}.We construct our hashtag corpus from ~\newcite{Wang2012},  and ~\newcite{Sintsova2017} \footnote{http://hci.epfl.ch/sharing-emotion-lexicons-and-data\#emo-hash-data}. More tweets between January and October 2017 are additionally added using the Twitter Firehose API by using the hashtags based on the hierarchy mentioned in ~\newcite{Shaver1987}. The hashtags are transformed into corresponding emotion labels of Joy, Sadness, Anger, and Fear.  When extending the dataset, we only use documents with emotional hashtags at the end and filter out any documents with URLs, quotations, or less than five words as ~\newcite{Wang2012} did. The total number of documents is about 1.9 million with four classes: joy (36.5\%), sadness (33.8\%), anger (23.5\%), and fear (6\%). The dataset is randomly split into a train (70\%), validation (15\%), and test set  (15\%) for experiments.

\subsubsection*{Small datasets}
For sentiment, we include 8 datasets. (1,2) SST-fine and SST-binary ~\cite{socher2013recursive} (3) OpeNER ~\cite{agerri2013opener} (4,5) tube\_auto and tube\_tablet ~\cite{uryupina2014sentube}  (6) SemEval \cite{hltcoe2013semeval} (7,8) SS-Twitter and SS-Youtube ~\cite{Thelwall2010}. For emotion tasks, we include 4 datasets, (1) ISEAR ~\cite{Wallbott1986} (2) WASSA ~\cite{Mohammad2017} (3) Olympic ~\newcite{Sintsova2013} (4) SE0714 ~\cite{staiano2014depechemood}. We further include 6 other affect-related datasets. (1,2) SCv1-GEN and SCv2-GEN for sarcasm detection, (3) Stress \cite{winata2018attention},  (4) Abusive ~\cite{waseem2016you,waseem2016hateful}. (5) Personality ~\cite{pennebaker1999linguistic} (6) Insult. 

\begin{table*}[t]
\centering

\resizebox{\textwidth}{!}{
\begin{tabular}{|c|cccccccc|c|}
\hline
model                   & SS-T & SS-Y & SS-binary & SS-fine & OpeNER & tube\_auto & tube\_tablet & SemEval & average \\ \hline
SSWE                    & \bf 0.815	& 0.835	& 0.698	& 0.365	& 0.701	& 0.620	& 0.654	& 0.629  & 0.665 \\ 
DeepMoji  embedding        & 0.788      & 0.841      & 0.751     & 0.369   & \bf 0.754  & 0.628      & 0.675        & \bf 0.676   & 0.685   \\ 
CNN embedding &  0.803      & \bf 0.862      & 0.734     & 0.369   & 0.713  & 0.605      & 0.667        & 0.622   & 0.672   \\ 
Emo2Vec                 & 0.801      & 0.859      & \bf 0.812     & \bf 0.416   & 0.744  & \bf 0.629      & \bf 0.688      & 0.638   & \bf 0.698   \\ \hline
\end{tabular}}
\vspace{-10pt}

\caption{Comparison between different emotion representations on sentiment datasets, all results are reported with accuracy. The best results are highlighted with bold fonts. Emo2Vec achieves best average score.}

\vspace{-10pt}
\label{table:comparison_sentiment}
\setlength{\abovecaptionskip}{5pt}
\end{table*}

\begin{table*}[t]
\centering

\resizebox{\textwidth}{!}{
\begin{tabular}{|c|cccccccccc|c|}
\hline
model              & ISEAR & WASSA & SE0714 & Olympic & Stress & SCv1-GEN & SCv2-GEN & Insult & Abusive & Personality & average \\ \hline
SSWE               & 0.327 & 0.466 & 0.217 & 0.508 & 0.704 & 0.660 & 0.678 & 0.559 & 0.539 & 0.674  & 0.533 \\ 
DeepMoji embedding     & 0.379 & 0.532 & 0.286  & 0.485   & 0.739  & 0.658    & 0.685    & \bf 0.666  & 0.586   & \bf 0.678   & 0.569     \\ 
CNN embedding & \bf 0.384 & 0.549 & 0.259  & 0.480   & \bf 0.744  & 0.657    & 0.707    & 0.623  & 0.560   & 0.676   & 0.564     \\ 
Emo2Vec            & 0.372 & \bf 0.559 & \bf 0.323  & \bf 0.506   & \bf 0.744  & \bf 0.674    & \bf 0.710    & 0.647  & \bf 0.588   & 0.675   & \bf 0.580     \\ \hline
\end{tabular}}
\vspace{-10pt}

\caption{Comparison between different representations on other affect related datasets. All results are reported with f1 score. The best results are highlighted with bold fonts. On average, Emo2Vec achieves best f1 score.}

\vspace{-10pt}
\label{table:comparison_others}
\end{table*}

\begin{table*}[t]
\centering

\resizebox{\textwidth}{!}{
\begin{tabular}{|c|cc|ccc|}
\hline
dataset      & \multicolumn{2}{c|}{Previous SOTA results}                                        & 
GloVe & GloVe+DeepMoji          & GloVe+Emo2Vec           \\ \hline
SS-Twitter   & bi-LSTM$~$\cite{Felbo2017}           & 0.88  & 0.78  & \bf 0.81  & \bf 0.81  \\
SS-Youtube   & bi-LSTM$~$\cite{Felbo2017}           & 0.93  & 0.84  &  0.86  & \bf 0.87  \\
SS-binary    & bi-LSTM$~$\cite{yu2017refining}      & 0.886 & 0.795 & 0.809                    & \bf 0.823 \\
SS-fine      & bi-LSTM$~$\cite{yu2017refining}      & 0.497 & 0.414 & 0.421                    & \bf 0.436 \\
OpeNER       & bi-LSTM$~$\cite{barnes2017assessing} & 0.825 & 0.750 & \bf 0.781 & 0.778                    \\
tube\_auto   & SVM$~$\cite{barnes2017assessing}     & 0.662 & 0.630 & 0.628                    & \bf 0.660 \\
tube\_tablet & SVM$~$\cite{barnes2017assessing}     & 0.681 & 0.650 &  0.678 &  \boxed{\textbf{0.684}} \\
SemEval      & bi-LSTM$~$\cite{barnes2017assessing} & 0.685 & 0.671 & \bf 0.695 & 0.680                    \\
ISEAR        & bi-LSTM$~$\cite{Felbo2017}           & 0.57  & 0.41  & 0.43                     & \bf 0.45  \\
SE0714       & bi-LSTM$~$\cite{Felbo2017}           & 0.37  & 0.36  & 0.36                     & \boxed{\textbf{0.43}}  \\
Olympic      & bi-LSTM$~$\cite{Felbo2017}           & 0.61  & 0.52  & 0.52                     & \bf 0.53  \\
stress       & bi-LSTM$~$\cite{winata2018attention}  & 0.743 & 0.759 & \bf 0.793 & 0.770                    \\
SCv1-GEN     & bi-LSTM$~$\cite{Felbo2017}           & 0.69  & \bf 0.69  &  0.68                     &  0.68  \\
SCv2-GEN     & bi-LSTM$~$\cite{Felbo2017}           & 0.75  & 0.73  & \bf 0.74                    & \bf 0.74  \\ \hline
Average      &                                                          &       & 0.642  & 0.657                    & \bf 0.667  \\ \hline
\end{tabular}
}
\vspace{-10pt}

\caption{Comparison between different word-level emotion representations with state-of-the-art results. The best results are in bold. New state-of-the-art results Emo2Vec that achieves are  highlighted with boxes.}

\vspace{-10pt}
\label{table:sota}
\end{table*}

\subsection{Pre-training Emo2Vec}
Emo2Vec embedding  matrix and the CNN model are pre-trained using hashtag corpus alone. Parameters of $T$ and CNN are randomly initialized and Adam is used for optimization. Best parameter settings are tuned on the validation set. For the best model, we use the batch size of 16, embedding size of 100, 1024 filters and filter sizes are 1,3,5 and 7 respectively. We keep the trained embedding and rename it as CNN embedding for comparison. 100-dim for Emo2Vec is used in all experiments.

\subsection{Multi-task training}
We tune our parameters of learning rate, L2 regularization, whether to pre-train our model and batch size with the average accuracy of the development set of all datasets. We early stop our model when the averaged dev accuracy stop increasing. Our best model uses learning rate of 0.001, L2 regularization of 1.0, batch size of 32. We save the best model and take the embedding layer as Emo2Vec vectors.  

\subsection{Evaluation}

\noindent\textbf{Baselines}: We use 50-dimension Sentiment-specific Word Embedding (SSWE) ~\cite{Tang2016} as our baseline, which is an embedding model  trained with 10 millions of tweets by encoding both semantic and sentiment information into vectors. Also, lots of work about the detection/classification in sentiment analysis implicitly encodes emotion inside the word vectors. For example, ~\newcite{Felbo2017} trains a two-layer bidirectional Long Short-Term Memory (bi-LSTM) model, named DeepMoji, to predict emoji of the input document using a huge dataset of 1.2 billion tweets. Their embedding layer is implicitly encoded with emotion knowledge. Thus, we use the DeepMoji embedding, the 256-dimension embedding layer of DeepMoji as another baseline. 





\noindent\textbf{Evaluation method}: To make a fair comparison with other baseline representations, we first take one dataset $d_i$ out from $n$ small datasets as the test set. The remaining $n-1$ small datasets and the larger dataset are used to train Emo2Vec through multi-task learning. We take the trained Emo2Vec as the feature for $d_i$ and train a logistic regression on $d_i$ to compare the performance with other baseline representations. The procedure is repeated $n$ times to see the generalization ability on different datasets. We release Emo2Vec trained on all datasets.  For sentiment tasks, accuracy score is reported. For other tasks,  if it is binary task, we report f1 score for the positive class. If it is multi-class classification tasks, we make it binary classification problem for each class and report averaged f1 score.


\section{Results}

We compare our Emo2Vec with SSWE, CNN embedding, DeepMoji embedding  and state-of-the-art(SOTA) results on 18 different datasets. The results can be found in Table \ref{table:comparison_sentiment} and 
Table \ref{table:comparison_others}.

\noindent\textbf{Compared with CNN embedding}: Emo2Vec works better than CNN embedding on 14/18 datasets,  giving 2.6\% absolute accuracy improvement for the sentiment task  and 1.6\% absolute f1-score improvement  on the other tasks. It shows multi-task training helps to create better generalized word emotion representations than just using a single task. 

\noindent\textbf{Compared with SSWE}: Emo2Vec works much better on all datasets except SS-T datasets, which gives 3.3\% accuracy improvement  and 4.7\% f1 score improvement respectively on sentiment and other tasks. This is because SSWE is trained on 10M binary classification task on twitter which then over-fits on dataset SS-T, and generalizes poorly to  other tasks. 

\noindent\textbf{Compared with DeepMoji embedding}: Emo2Vec outperforms DeepMoji on 13/18 datasets despite the much smaller size of our training corpus (1.9M documents for us vs 1.2B documents for DeepMoji). On average, it gives 1.3\% improvement in accuracy for the sentiment task  and 1.1\% improvement of f1-score on the other tasks. 

\noindent\textbf{Compared with SOTA results}: We further compare the performance of Emo2Vec vectors with SOTA results on 14 datasets where the same split is shared. Since Emo2Vec is not trained by predicting contextual words, it is weak on capturing synthetic and semantic meaning. Thus, we concatenate Emo2Vec with the pre-trained GloVe vectors, which are trained on Twitter and Wikepedia \footnote{http://nlp.stanford.edu/data/glove.twitter.27B.zip and http://nlp.stanford.edu/data/glove.6B.zip}. Then, the concatenated vector of GloVe and Emo2Vec, the concatenated vector of GloVe and DeepMoji embeddings and GloVe  are included for comparison with SOTA results. Note that SOTA results require complex bi-LSTM model while all these representations are trained and reported with a logistic regression classifier. Here, we want to highlight that solely using a simple classifier with good word representation can achieve promising results.

Table \ref{table:sota} shows that GloVe+Emo2Vec outperforms GloVe on 13/14 datasets. Compared with GloVe+DeepMoji, GloVe+Emo2Vec achieves same or better results on 11/14 datasets, which on average gives 1.0\% improvement.  GloVe+Emo2Vec achieves better performances on SOTA results on three datasets (SE0714, stress and tube\_tablet) and comparable result to SOTA on another four datasets (tube\_auto, SemEval, SCv1-GEN and SCv2-GEN). We believe the reason why we achieve a much better performance than SOTA  on the SE0714 is that headlines are usually short and emotional words exist more commonly in headlines. Thus, to detect the corresponding emotion, more attention needs to be paid to words. 

\section{Related work}

For sentiment analysis, numerous classification models ~\cite{kalchbrennerconvolutional, iyyer2015deep, dou2017capturing} have been explored. Multi-modal sentiment analysis ~\cite{zadeh2017tensor,poria2017review} extends text-based model to the combination of visual, acoustic and language, which achieves better results than the single modality. Various methods are developed for automatic constructions of sentiment lexicons using both supervised and unsupervised way~\cite{wang2017sentiment}.  Aspect-based sentiment ~\cite{chen2017recurrent,wang2016attention} is also a hot topic where researchers care more about the sentiment towards a certain target. Transfer learning from the large corpus is also investigated by ~\newcite{Felbo2017} to train a large model on a huge emoji tweet corpus, which boosts the performance of affect-related tasks. Multi-task training has achieved great success in various natural language tasks, such as machine translation~\cite{dong2015multi, malaviya2017learning}, multilingual tasks~\cite{duong2015low,gillick2016multilingual}, semantic parsing~\cite{peng2017deep}. \newcite{hashimoto2017joint} jointly learns POS tagging, chunking, dependency parsing, semantic relatedness, and textual entailment by considering linguistic hierarchy and achieves state-of-the-results on five datasets.  For sentiment analysis, ~\newcite{Balikas2017multi} jointly trains ternary and fine-grained classification with a recurrent neural network and achieves new  state-of-the-art results.

\section{Conclusion and Future Work}
%

In this paper, we propose Emo2Vec to represent emotion with vectors using a multi-task training framework. Six affect-related tasks are utilized, including  emotion/sentiment analysis, sarcasm classification, stress detection, abusive language classification, insult detection, and personality recognition. We empirically show how Emo2Vec leverages multi-task training to learn a generalized emotion representation. In addition, Emo2Vec outperforms existing affect-related embeddings on more than ten different datasets. By combining Emo2Vec with GloVe, logistic regression can achieve competitive performances on several state-of-the-art results.

\section{Acknowledgements}

This work is partially funded by ITS/319/16FP of Innovation Technology Commission, HKUST 16248016 of Hong Kong Research Grants Council.





\bibliography{emnlp2018}
\bibliographystyle{acl_natbib_nourl}

\appendix

\newpage
\newpage
\clearpage

\end{document}